\documentclass[10pt,twocolumn,letterpaper]{article}

\usepackage[pagenumbers]{cvpr} 

\newcommand{\myparagraph}[1]{\vspace{1.5ex}\noindent\textbf{#1}}

\definecolor{cvprblue}{rgb}{0.21,0.49,0.74}
\usepackage[pagebackref,breaklinks,colorlinks,allcolors=cvprblue]{hyperref}
\usepackage[accsupp]{axessibility}

\usepackage{multirow}
\usepackage{pifont}
\usepackage{amsmath}
\usepackage{amssymb}
\usepackage{booktabs}
\usepackage{graphicx}
\newcommand{\cmark}{\ding{51}}%
\newcommand{\xmark}{\ding{55}}%

\usepackage[ruled,vlined]{algorithm2e}

\def\paperID{} 
\def\confName{CVPR}
\def\confYear{2026}

\title{CIGPose: Causal Intervention Graph Neural Network for Whole-Body Pose Estimation}

\author{
    Bohao Li\textsuperscript{1} \quad
    Zhicheng Cao\textsuperscript{2}\thanks{Corresponding authors.} \quad
    Huixian Li\textsuperscript{1}\footnotemark[1] \quad
    Yangming Guo\textsuperscript{3}\footnotemark[1] \\
    \textsuperscript{1}School of Computer Science, Northwestern Polytechnical University \\
    \textsuperscript{2}Xidian University \\
    \textsuperscript{3}School of Cybersecurity, Northwestern Polytechnical University \\
    {\tt\small bh\_li@mail.nwpu.edu.cn, zccao@xidian.edu.cn, \{lihuixian, yangming\_g\}@nwpu.edu.cn}
}

\begin{document}
\maketitle
\begin{abstract}
State-of-the-art whole-body pose estimators often lack robustness, producing anatomically implausible predictions in challenging scenes. We posit this failure stems from spurious correlations learned from visual context, a problem we formalize using a Structural Causal Model (SCM). The SCM identifies visual context as a confounder that creates a non-causal backdoor path, corrupting the model's reasoning. We introduce the Causal Intervention Graph Pose (CIGPose) framework to address this by approximating the true causal effect between visual evidence and pose. The core of CIGPose is a novel Causal Intervention Module: it first identifies confounded keypoint representations via predictive uncertainty and then replaces them with learned, context-invariant canonical embeddings. These deconfounded embeddings are processed by a hierarchical graph neural network that reasons over the human skeleton at both local and global semantic levels to enforce anatomical plausibility. Extensive experiments show CIGPose achieves a new state-of-the-art on COCO-WholeBody. Notably, our CIGPose-x model achieves 67.0\% AP, surpassing prior methods that rely on extra training data. With the additional UBody dataset, CIGPose-x is further boosted to 67.5\% AP, demonstrating superior robustness and data efficiency. The codes and models are publicly available at \url{https://github.com/53mins/CIGPose}.
\end{abstract}    
\section{Introduction}
\label{sec:intro}

\begin{figure}[t]
    \centering
    \includegraphics[width=\linewidth]{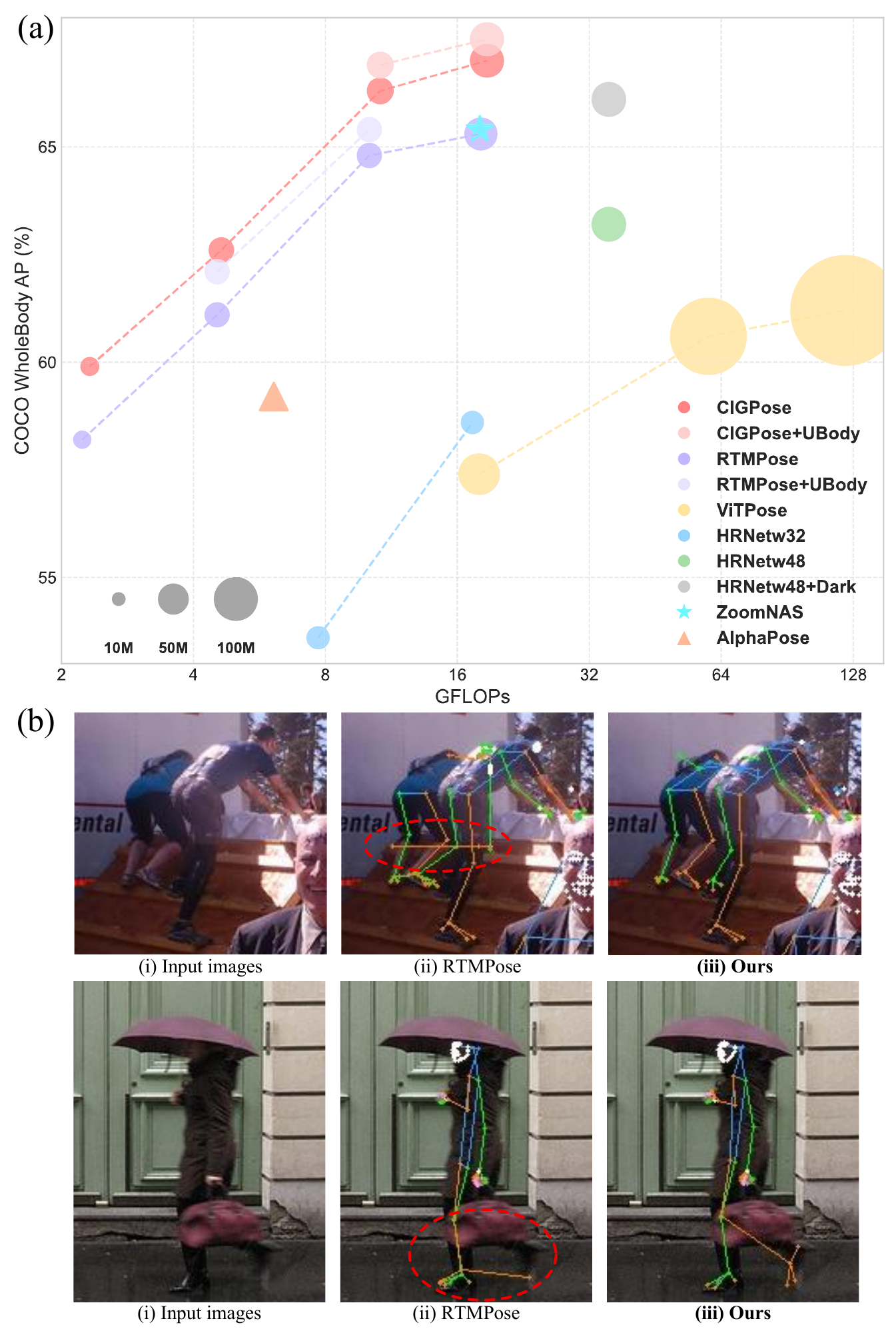}
    \caption{(a) Comparison of CIGPose with related models for whole-body pose estimation on COCO-WholeBody. (b) Qualitative comparison between CIGPose-x and RTMPose-x \cite{jiang2023rtmpose}.}
    \label{fig:result_contrast}
    \vspace{-1em}
\end{figure}

Whole-body pose estimation—localizing dense anatomical keypoints \cite{cao2019openpose,hidalgo2019single,jin2020whole} for the entire human body, including the limbs, face, and hands—is fundamental for applications like 3D mesh recovery \cite{lin2023one,wang2025blade,li2025hybrik}, motion generation \cite{yi2023generating,lin2023motion}, and human-robot interaction \cite{sampieri2022pose,fan2023arctic}. Despite deep learning's progress, state-of-the-art models \cite{jiang2023rtmpose,yang2023effective,jiang2024rtmw} lack robustness, often failing in real-world scenarios with heavy occlusion, clutter, or difficult lighting. This fragility suggests high-capacity models \cite{xu2023vitpose++,khirodkar2024sapiens} rely on superficial statistics rather than anatomical understanding.

We posit the root cause is visual confounding \cite{liu2022towards}, leading to failures like misidentifying background patterns as limbs, as shown in \cref{fig:result_contrast}(b). These errors stem from spurious correlations; for example, a network might associate a "backrest" with a "torso" due to co-occurrence in training data. This context $C$ confounds the extracted features $F$ and the final pose $Y$, creating a non-causal backdoor path $F \leftarrow X \leftarrow C \to Y$ that corrupts the model's prediction, which is based on the observational distribution $P(Y|F)$. True robustness requires learning the interventional distribution $P(Y|do(F))$, using the $do$-operator \cite{pearl2016causal} to remove the confounder's influence.

We introduce the \textbf{C}ausal \textbf{I}ntervention \textbf{G}raph Neural Network for Whole-body Pose Estimation (CIGPose), a framework approximating the causal intervention $P(Y|do(F))$. Our core contribution is a Causal Intervention Module (CIM) that identifies confounded keypoint embeddings using predictive uncertainty as a proxy. The CIM then replaces these corrupted embeddings with learned, context-invariant canonical embeddings, blocking the backdoor path and forcing the model to use causally sound evidence.

A hierarchical graph neural network then processes these deconfounded embeddings to enforce global anatomical plausibility. The GNN models the human skeleton at multiple semantic levels, first local kinematics, then long-range dependencies, enforcing structural consistency on a causal foundation to produce accurate and anatomically coherent poses.

CIGPose establishes a new state-of-the-art on COCO-WholeBody \cite{jin2020whole}, as shown in \cref{fig:result_contrast}(a), and its effectiveness is also assessed on COCO \cite{lin2014microsoft} and CrowdPose \cite{li2019crowdpose} datasets. Trained only on COCO-WholeBody \cite{jin2020whole}, our CIGPose-x model reaches 67.0\% AP, outperforming competitors that use additional datasets. When trained with the additional UBody \cite{lin2023one} dataset, its performance rises to 67.5\% AP, surpassing 66.5\% AP of DWPose-l \cite{yang2023effective} and demonstrating superior robustness and data efficiency.

In summary, our main contributions are:
\begin{itemize}
    \item We formalize 2D whole-body pose estimation within a causal framework, identifying visual context as a critical confounder that creates spurious correlations.
    \item We propose a novel Causal Intervention Module (CIM) that approximates a causal $do$-operation by identifying and replacing confounded keypoint embeddings with learned canonical representations.
    \item We introduce a hierarchical graph neural network that explicitly models anatomical structure on deconfounded embeddings, enhancing global pose consistency.
    \item CIGPose achieves state-of-the-art performance on multiple large-scale benchmarks, demonstrating superior accuracy and robustness in complex scenes.
\end{itemize}
\section{Related Work}

\subsection{2D Whole-body Pose Estimation}

Modern top-down 2D pose estimation \cite{jin2020whole,xu2022zoomnas} evolved from high-resolution CNNs \cite{sun2019deep} to Vision Transformers \cite{yuan2021hrformer,xu2022vitpose,xu2023vitpose++}, with a concurrent focus on efficiency \cite{jiang2023rtmpose,yang2023effective}. Yet, SOTA estimators fail in complex scenes due to spurious correlations learned from visual context. Prior work improved robustness via knowledge distillation \cite{yang2023effective} or massive datasets \cite{jiang2024rtmw,khirodkar2024sapiens} but did not address confounding directly. Our causal framework achieves inherent robustness by identifying and mitigating visual confounders, improving both accuracy and data efficiency. Other works address specific robustness challenges. ProbPose \cite{purkrabek2025probpose} for out-of-image keypoints and robustness for 2D-to-3D lifters \cite{hoang2024improving}. Video-based whole-body motion estimation also explores robustness, such as REWIND \cite{lee2025rewind}.

\subsection{Graph Neural Networks for Human Pose Estimation}

The skeleton's structure is a natural fit for GNNs, used to enforce anatomical constraints by modeling joint kinematics \cite{zhao2019semantic,shi2019skeleton,peng2020learning,hu2021conditional}.
Hierarchical GNNs \cite{zeng2021learning,quan2021higher,li2021hierarchical,lee2023hierarchically,zheng2025hipart}, inspired by action recognition, refine poses by reasoning at multiple semantic levels, from local limbs to global inter-part relations. Yet, these methods \cite{wang2020graph,wang2021robust,zhang20233d,azizi2024occlusion} are vulnerable to confounded inputs; a GNN will propagate errors from corrupted initial keypoint representations \cite{jin2020differentiable}. Many are also designed for video \cite{yang2021learning,zeng2021learning,yu2023gla,he2024video}, integrating temporal information, unlike our single-image task. Our framework is to address this issue of error propagation. A causal intervention module precedes the GNN, ensuring structural reasoning occurs on a deconfounded embedding set, making the model more robust.

\begin{figure*}[t]
    \centering
    \includegraphics[width=\linewidth]{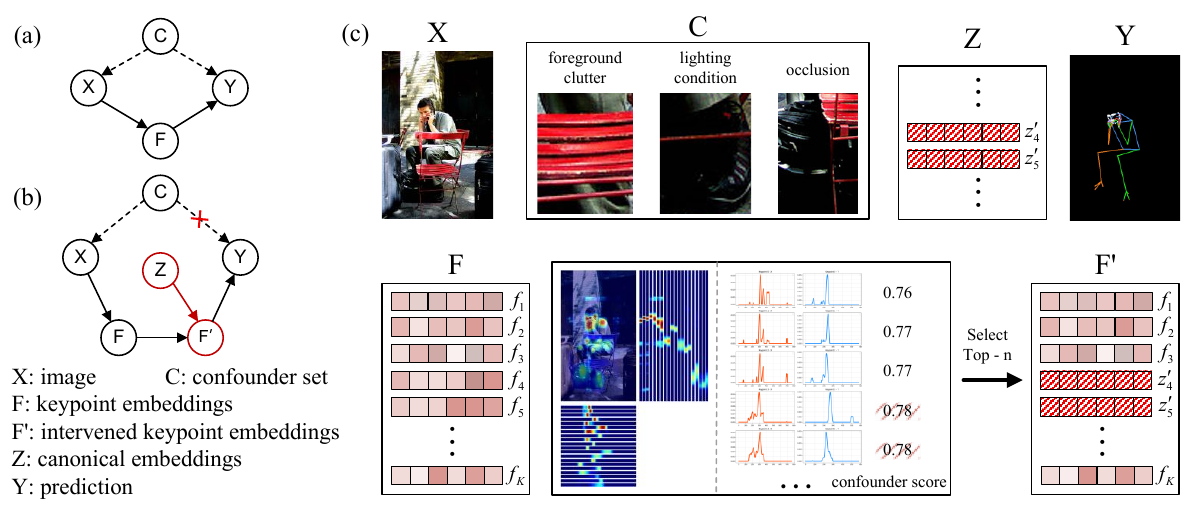}
    \caption{(a) The proposed Structural Causal Model (SCM) for keypoint estimation, (b) The intervened SCM after applying the $do$-operator to the keypoint embeddings, (c) The realization of each component within our Causal Intervention Module (CIM).}
    \label{fig:cim}
    \vspace{-1em}
\end{figure*}

\subsection{Causal Inference}

Causal inference \cite{pearl2014interpretation} improves robustness by learning the interventional distribution $P(Y|do(X))$ instead of the observational $P(Y|X)$. In computer vision, scene context is a known confounder creating a backdoor path, formalized with Structural Causal Models (SCMs) \cite{pearl2009causality} in tasks like semantic segmentation \cite{zhang2020causal} and VQA \cite{niu2021counterfactual}. The backdoor adjustment formula is intractable in high-dimensional vision, leading to practical approximations \cite{yue2021causal,su2021context,israel2023high} like marginalizing out the context. Related ideas of explicit representation decoupling have also been explored beyond pose estimation, for example in hyperspectral image super-resolution and pansharpening \cite{du2026unsupervised,du2026pansharpening}. Our work introduces a novel mechanism: we identify confounded embeddings via predictive uncertainty and replace them with learned, context-invariant canonical embeddings, severing the backdoor path. Notably, CleanPose \cite{lin2025cleanpose} addresses data biases by integrating causal learning with knowledge distillation; it employs a front-door adjustment, distinct from our intervention strategy.
\section{Method}
\label{sec:method}

\subsection{Structural Causal Model and Theoretical Goal}

We formalize pose estimation using the Structural Causal Model (SCM) \cite{pearl2009causality} in \cref{fig:cim}(a), defining relationships between the image $X$, unobserved confounders $C$, keypoint embeddings $F$, and prediction $Y$.

\begin{itemize}
    \item $C \to X$: Confounder set $C$, such as specific lighting conditions or foreground clutter, is inherent to the captured scene and thus causally determines the visual content of the input image $X$.
    \item $X \to F$: A keypoint encoder processes the input image $X$ to produce a set of initial keypoint embeddings $F$.
    \item $C \to Y$: This path shows the core confounding problem: The model learns a spurious, non-causal correlation from the training data between visual context $C$ (\eg "backrest") and true pose $Y$ (\eg a "torso" in a sitting position), thus mistaking context for evidence.
    \item $F \to Y$: The prediction head decodes the keypoint embeddings $F$ to produce the final coordinate predictions $Y$. This represents the true, desired causal path.
\end{itemize}

The critical issue is the non-causal backdoor path $F \leftarrow X \leftarrow C \to Y$. This path allows the model to exploit spurious correlations: the feature embeddings $F$ (which are caused by $X$, which is caused by $C$) become associated with $Y$, not through the desired causal path $F \to Y$, but via the confounding arc $C \to Y$. A model trained on this confounded data learns the observational distribution $P(Y|F)$, not the true causal relationship.

To eliminate this confounding effect, our goal is to estimate the interventional distribution $P(Y|do(F))$. The $do$-operator \cite{pearl2016causal} signifies this intervention. The intervention aims to block the non-causal backdoor path $F \leftarrow X \leftarrow C \to Y$. As shown in \cref{fig:cim}(b), this is conceptually equivalent to severing the confounding link $C \to Y$, forcing the model to rely only on the causal path $F \to Y$. Theoretically, this causal effect can be computed using the backdoor adjustment formula:
\begin{equation}
\label{eq:backdoor_adjustment}
P(Y|do(F)) = \sum_{c} P(Y|F,c)P(c)
\end{equation}

However, this formula is intractable in our scenario. The confounder $C$ represents all possible high-dimensional visual contexts and is unobserved, making the summation over $c$ impossible to compute. Therefore, we propose the \textbf{C}ausal \textbf{I}ntervention \textbf{M}odule (CIM), a novel and practical framework designed to approximate this causal intervention. We provide a formal derivation for \cref{eq:backdoor_adjustment} using Pearl's $do$-calculus \cite{pearl2016causal} and a theoretical justification for our counterfactual approximation in Appendix \ref{sec:appendix_theory}.

\subsection{Causal Intervention via Counterfactual Replacement}

Our CIM approximates \cref{eq:backdoor_adjustment} by identifying confounded embeddings $F_{\mathit{conf}} \subset F$ and replacing them with a "clean", context-invariant representation.

\textbf{Justification for Counterfactual Replacement.}
We propose to replace a confounded embedding $f_k \in F_{\mathit{conf}}$ with a learned canonical embedding $z_k$. These are drawn from a learnable embedding table $Z \in \mathbb{R}^{K \times d_{\mathit{emb}}}$, optimized end-to-end to represent a context-invariant ideal for each keypoint type. The key insight is that $Z$, as a set of learnable model parameters, is by construction independent of the specific confounder instance $C$ affecting any given input image, such that $Z \perp C$. By performing this counterfactual intervention $do(f_k := z_k)$, we replace the confounded embedding with an embedding that is independent of $C$. This action breaks the dependency chain $F \leftarrow X \leftarrow C$, thereby blocking the non-causal backdoor path $F \leftarrow X \leftarrow C \to Y$.

\textbf{Justification for Confounder Identification.}
To intervene selectively, we must identify $F_{\mathit{conf}}$. We posit that predictive uncertainty is an effective proxy for confounding. The confounder set $C$ creates a conflict between visual evidence $F$ and biased priors (from $C \to Y$), manifesting as high epistemic uncertainty. While this hypothesis applies generally, validating it against ambiguous, hard-to-label factors like clutter or contextual confusion is intractable. However, we can robustly validate our proxy against heavy occlusion, which is arguably the most severe and clearly-defined class of confounder in pose estimation.

\begin{figure}[htbp]
\centering
\begin{subfigure}[b]{0.49\linewidth}
    \includegraphics[width=\textwidth]{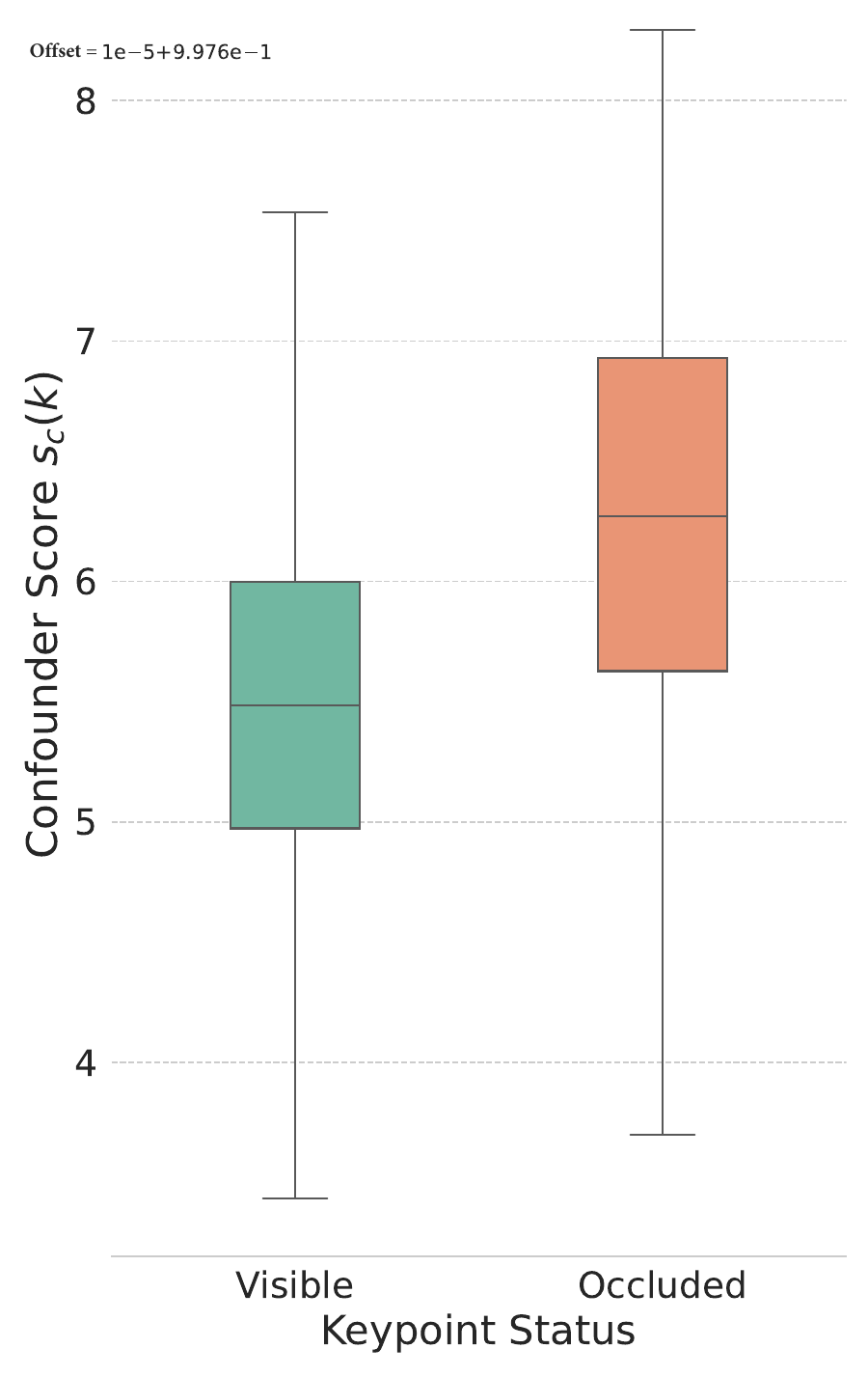}
    \caption{Limb Keypoints}
    \label{fig:confounder_boxplot_limbs}
\end{subfigure}
\hfill
\begin{subfigure}[b]{0.49\linewidth}
    \includegraphics[width=\textwidth]{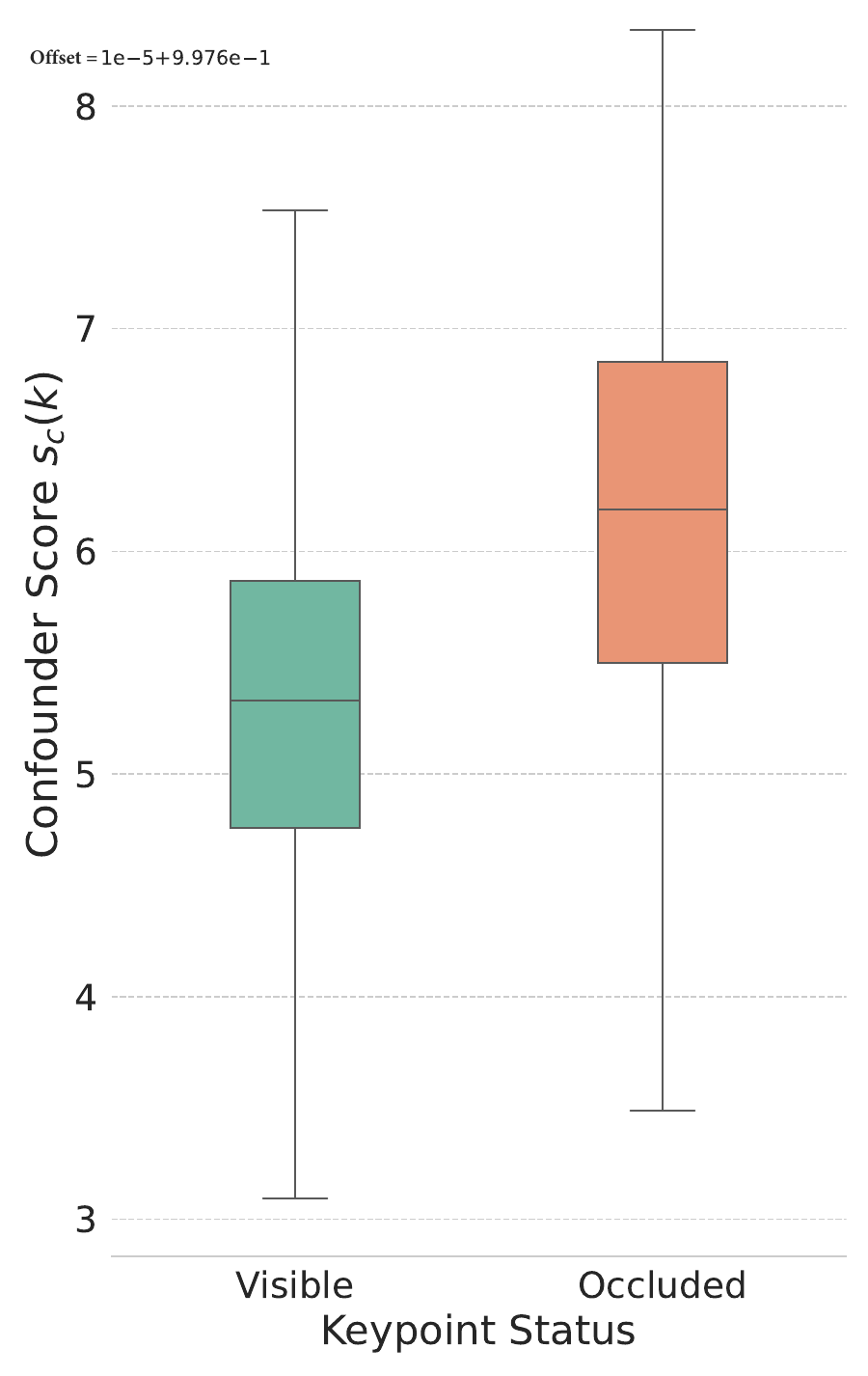}
    \caption{All Keypoints}
    \label{fig:confounder_boxplot_all}
\end{subfigure}
\caption{Validation of confounder score $s_c(k)$ on COCO-WholeBody \cite{jin2020whole}. The confounder score is significantly higher for occluded keypoints than for visible ones, for both (a) easily occluded limb keypoints and (b) all keypoints combined.}
\label{fig:confounder_boxplot}
\end{figure}

High uncertainty is reflected in diffuse posterior probability distributions (see Appendix \ref{sec:appendix_uncertainty_viz}). We quantitatively validate this in \cref{fig:confounder_boxplot}. This figure plots our proposed confounder score $s_c(k)$ (defined in \cref{eq:confounder_identification}) against the keypoint's ground-truth occlusion status on the COCO-WholeBody \cite{jin2020whole} validation set. As shown in \cref{fig:confounder_boxplot}(a) and \cref{fig:confounder_boxplot}(b), occluded keypoints exhibit a significantly higher median and wider distribution of confounder scores compared to visible keypoints. This positive correlation supports our assumption that predictive uncertainty is a reliable proxy for detecting this critical type of confounding. We hypothesize this sensitivity to occlusion-induced ambiguity also grants robustness to other non-occlusion-based confounders, supported by our model's performance in complex scenes as shown in \cref{fig:contrast}. Further details on the network architecture of CIM are available in Appendix \ref{sec:appendix_implementation}.

\textbf{Implementation Details.}
Based on this justification, the CIM (illustrated in \cref{fig:cim}(c)) operates via a two-step process:

\textbf{Step 1. Confounder Identification.} Given the initial keypoint embeddings $F$, we first generate 1D coordinate heatmaps, normalized into posterior probability distributions $(P_{k,x}, P_{k,y})$. We use the concentration of this distribution to quantify predictive ambiguity. A low peak signals high uncertainty, which we use as a proxy for confounding. We define the confounder score for each keypoint as:
\begin{equation}
\label{eq:confounder_identification}
s_c(k) = 1 - \frac{1}{2}(\max(P_{k,x}) + \max(P_{k,y}))
\end{equation}
The keypoints with the top-$n$ highest scores $s_c(k)$ are selected for intervention.

\textbf{Step 2. Counterfactual Embedding Replacement.} For each selected embedding $f_k \in F$, we perform the intervention $do(f_k := z_k)$ by replacing it with its corresponding canonical embedding $z_k$ from the learnable table $Z$. This substitution creates the deconfounded, or "cleaned", embedding set $F'$:
\begin{equation}
\label{eq:cf_embedding_replace}
f'_{k} = \begin{cases} z_k, & \text{if } k \text{ is selected for intervention} \\ f_k, & \text{otherwise} \end{cases}
\end{equation}

The final prediction $Y$ is then derived from this cleaned set $F'$. By replacing confounded embeddings with their ideal counterparts, we block the spurious causal path from $C$ and force the model to reason from a more robust, causally sound representation.

\begin{figure}[!htbp]
    \centering
    \includegraphics[width=0.88\linewidth]{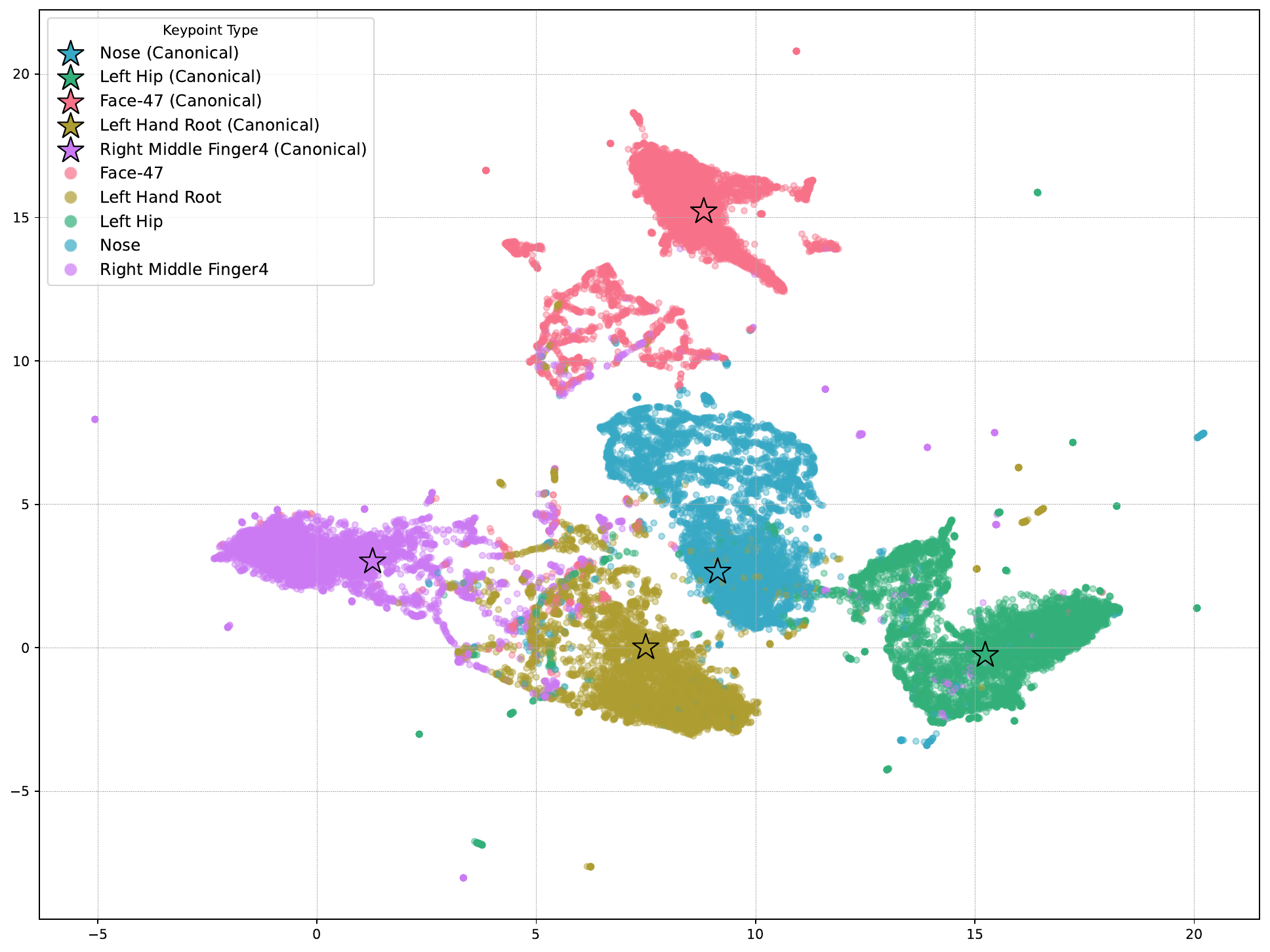}
    \caption{UMAP visualization of initial contextual embeddings vs. their corresponding learned canonical embeddings.}
    \label{fig:umap}
\end{figure}

\textbf{Analysis of Canonical Embeddings.}
To validate our assumption that $Z$ learns context-invariant ideals, we visualize the embeddings using UMAP in \cref{fig:umap}. The initial contextual embeddings $F$ (from the encoder) for a single keypoint type form a diffuse cluster, reflecting variance from visual confounders. In contrast, the corresponding learned canonical embedding $z_k$ (from $Z$) is learned as a single, highly concentrated point. This demonstrates that our model successfully learns a stable, context-invariant representation for each keypoint, which serves as the "causal ideal" during intervention. The clear separation between canonical embeddings for different keypoint types further indicates their semantic distinctiveness.

\subsection{Hierarchical Graph Reasoning on Deconfounded Embeddings}
\label{sec:hierarchical_gnn}

\begin{figure}[htbp]
    \centering
    \includegraphics[width=\linewidth]{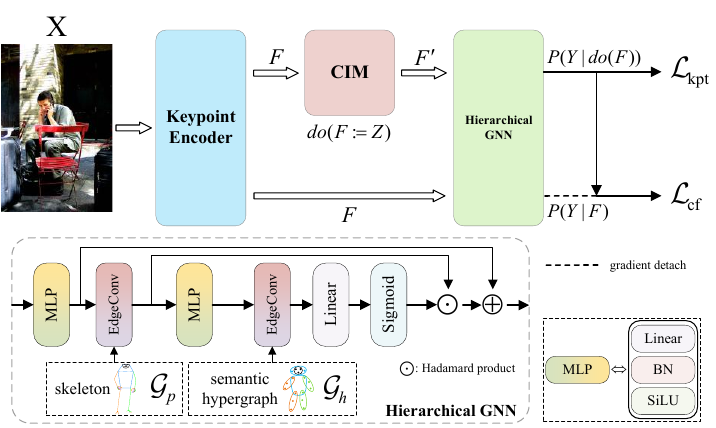}
    \caption{Overview of our CIGPose architecture. During training, embeddings are processed in two paths: (1) a counterfactual path where our CIM module deconfounds them before the Hierarchical GNN, and (2) an observational path using original embeddings for consistency. Inference relies solely on the counterfactual path.}
    \label{fig:pipeline_structure}
\end{figure}

To enforce global anatomical constraints on the purified embeddings $F'$, we employ a two-stage hierarchical graph neural network (GNN). Our design, inspired by HD-GCN \cite{lee2023hierarchically}, first models local anatomical regions before reasoning about their global inter-relationships, as shown in \cref{fig:pipeline_structure}.

\textbf{Intra-Part Relational Modeling.} The first stage uses an EdgeConv \cite{wang2019dynamic} layer over the standard anatomical skeleton graph $\mathcal{G}_p = (\mathcal{V}, \mathcal{E}_p)$, where edges $\mathcal{E}_p$ represent physical connections. This models local kinematic relationships by updating each keypoint embedding based on its neighbors.

\textbf{Inter-Part Contextual Attention.} The second stage captures long-range dependencies using a semantic hypergraph, denoted as $\mathcal{G}_h$. We predefine hyperedges $\mathcal{E}_h$ as functional keypoint groups (\eg 'left hand'). A representation for each hyperedge $g_e$ is computed by aggregating its constituent keypoint embeddings: $g_e = \frac{1}{|e|} \sum_{k \in e} f'_k$. These hyperedge representations then exchange information via message-passing to become context-aware. Finally, these context-aware representations generate channel-wise attention weights to modulate and refine the keypoint embeddings, producing the final, anatomically coherent embeddings $F''$:
\begin{equation}
\label{eq:attention}
f''_{k} = f'_{k} \odot \left( \frac{1}{|\mathcal{E}_k|} \sum_{e \in \mathcal{E}_k} \sigma(\psi_a(g'_e)) \right)
\end{equation}
where $g'_e$ is the updated, context-aware representation for a hyperedge $e$, $\psi_a$ is an MLP, $\sigma$ is the sigmoid function, and $\mathcal{E}_k$ is the set of hyperedges containing keypoint $k$. This hierarchical process produces embeddings that are both deconfounded and anatomically plausible. We provide expanded implementation details for the hierarchical GNN components in Appendix \ref{sec:appendix_implementation}.

\subsection{Joint Optimization with Counterfactual Consistency}

CIGPose is trained end-to-end with a composite objective that combines a primary supervised loss with a regularizer to ensure our causal interventions are meaningful (see \cref{fig:pipeline_structure}).

The primary prediction loss, $\mathcal{L}_{\text{kpt}}$, is applied to the final output of the counterfactual path. It minimizes the KL divergence between the final predicted pose distribution $P(Y|do(F))$ and the ground-truth distribution $Q$:
\begin{equation}
\label{eq:loss_kpt}
\mathcal{L}_{\text{kpt}} = \sum_{k=1}^{K} w_k \cdot D_{KL}(Q_k \mid\mid P(Y_k|do(F)))
\end{equation}
where $w_k$ weights keypoints by their ground-truth visibility.

To ensure interventions are targeted and canonical embeddings $Z$ are meaningful, we introduce a Counterfactual Consistency Loss, $\mathcal{L}_{\text{cf}}$. It regularizes the model by penalizing the divergence between the prediction from the observational path, $P(Y|F)$, and the prediction from the counterfactual path, $P(Y|do(F))$. This loss is only applied to the set of stable keypoints $S$, which is defined as the $K-n$ keypoints with the lowest confounder scores $s_c(k)$ that were not selected for intervention in \cref{eq:cf_embedding_replace}.
\begin{equation}
\label{eq:loss_cf}
\mathcal{L}_{\text{cf}} = \frac{1}{|S|} \sum_{k \in S} D_{KL}( \text{sg}[P(Y_k|F)]) \mid\mid P(Y_k|do(F))
\end{equation}
where the stop-gradient operator, $\text{sg}[\cdot]$, treats the observational prediction as a fixed target. This encourages the intervention to only alter confounded representations without disrupting reliable ones. The final training objective is a weighted sum of these two components: $\mathcal{L} = \mathcal{L}_{\text{kpt}} + \lambda \mathcal{L}_{\text{cf}}$, where $\lambda$ is a weight hyperparameter.
\section{Experiments}
\label{sec:exp}

\subsection{Datasets and Evaluation Metrics}

\begin{table*}[!htb]
\begin{center}
\setlength{\tabcolsep}{4.25pt}
\caption{Whole-body pose estimation results on COCO-WholeBody \cite{jin2020whole,xu2022zoomnas} V1.0 dataset. ``*'' denotes the model that relies on two-stage distillation and additional training data from the UBody dataset \cite{lin2023one}. ``\dag'' indicates multi-scale testing. Flip test is used.}\label{tab:compare_cocowhole}
		\begin{tabular}{c|l|c|c|cc|cc|cc|cc|cc}
			\toprule
			  & Method & Input Size & GFLOPs & \multicolumn{2}{c|}{whole-body} & \multicolumn{2}{c|}{body}  & \multicolumn{2}{c|}{foot}  & \multicolumn{2}{c|}{face}  & \multicolumn{2}{c}{hand} \\
			\cmidrule{5-14}
			& &   &   &  AP     & AR     & AP   & AR     &  AP  & AR     & AP    & AR   &  AP     & AR  \\
			\midrule
			Whole- & SN\dag~\cite{hidalgo2019single} & N/A & 272.3 & 32.7 & 45.6 & 42.7 & 58.3 & 9.9 & 36.9 & 64.9 & 69.7 & 40.8 & 58.0  \\ 
            body & OpenPose~\cite{cao2019openpose} & N/A & 451.1 & 44.2 & 52.3 & 56.3 & 61.2 & 53.2 & 64.5 & 76.5 & 84.0 & 38.6 & 43.3  \\ 
            \midrule
			Bottom- & PAF\dag~\cite{cao2017realtime} & 512$\times$512 & 329.1 & 29.5 & 40.5 & 38.1 & 52.6 & 5.3 & 27.8 & 65.6 & 70.1 & 35.9 & 52.8  \\ 
			up & AE~\cite{newell2017associative} & 512$\times$512 & 212.4 & 44.0 & 54.5 & 58.0 & 66.1 & 57.7 & 72.5 & 58.8 & 65.4 & 48.1 & 57.4  \\
			\midrule
            & SimpleBaseline~\cite{xiao2018simple} & 384$\times$288 & 20.4 & 57.3 & 67.1 & 66.6 & 74.7 & 63.5 & 76.3 & 73.2 & 81.2 & 53.7 & 64.7 \\
			& HRNet~\cite{sun2019deep}  & 384$\times$288 & 16.0 & 58.6 & 67.4 & 70.1 & 77.3 & 58.6 & 69.2 & 72.7 & 78.3 & 51.6 & 60.4  \\
            & PVT~\cite{wang2021pyramid} & 384$\times$288 & 19.7 & 58.9 & 68.9 & 67.3 & 76.1 & 66.0 & 79.4 & 74.5 & 82.2 & 54.5 & 65.4 \\
		    & FastPose50-dcn-si~\cite{fang2022alphapose} &  256$\times$192 & 6.1 & 59.2 & 66.5 & 70.6      & 75.6 & 70.2 & 77.5 & 77.5 & 82.5 & 45.7 & 53.9 \\
                & ZoomNet~\cite{jin2020whole} & 384$\times$288 & 28.5 & 63.0 & 74.2 & 74.5 & 81.0 & 60.9 & 70.8 & 88.0 & 92.4 & 57.9 & 73.4   \\
            & ZoomNAS~\cite{xu2022zoomnas} & 384$\times$288 & 18.0 & 65.4 & 74.4 & 74.0 & 80.7 & 61.7 & 71.8 & 88.9 & 93.0 & 62.5 & 74.0 \\
            & ViTPose+-H~\cite{xu2023vitpose++} & 256$\times$192 & 122.9 & 61.2 & - & 75.9 & - & 77.9 & - & 63.3 & - & 54.7 & -   \\
            & RTMPose-m~\cite{jiang2023rtmpose} & 256$\times$192 & 2.2 & 58.2 & 67.4 & 67.3 & 75.0 & 61.5 & 75.2 & 81.3 & 87.1 & 47.5 & 58.9  \\
            & RTMPose-l~\cite{jiang2023rtmpose} & 256$\times$192 & 4.5 & 61.1 & 70.0 & 69.5 & 76.9 & 65.8 & 78.5 & 83.3 & 88.7 & 51.9 & 62.8  \\
            Top-& RTMPose-l~\cite{jiang2023rtmpose} & 384$\times$288 & 10.1 & 64.8 & 73.0 & 71.2 & 78.1 & 69.3 & 81.1 & 88.2 & 91.9 & 57.9 & 67.7 \\
            down & RTMPose-x~\cite{jiang2023rtmpose} & 384$\times$288 & 18.1 & 65.3 & 73.3 & 71.4 & 78.4 & 69.2 & 81.0 & 88.9 & 92.3 & 59.0 & 68.5 \\
            \cmidrule{2-14}
            & RTMPose-l + UBody & 256$\times$192 & 4.5 & 62.1 & 70.6 & 69.7 & 76.9 & 65.5 & 78.1 & 84.1 & 89.3 & 55.1 & 65.4 \\
            & RTMPose-l + UBody & 384$\times$288 & 10.1 & 65.4 & 73.2 & 71.0 & 77.9 & 68.6 & 80.2 & 88.5 & 92.2 & 60.6 & 69.9 \\
            & DWPose-m*~\cite{yang2023effective} & 256$\times$192 & 2.2 & 60.6 & 69.5 & 68.5 & 76.1 & 63.6 & 77.2 & 82.8 & 88.1 & 52.7 & 63.4 \\
            & DWPose-l*~\cite{yang2023effective} & 256$\times$192 & 4.5 & 63.1 & 71.7 & 70.4 & 77.7 & 66.2 & 79.0 & 84.3 & 89.4 & 56.6 & 66.5 \\
            & DWPose-l*~\cite{yang2023effective} & 384$\times$288 & 10.1 & 66.5 & 74.3 & 72.2 & 78.9 & 70.4 & 81.7 & 88.7 & 92.1 & 62.1 & 71.0 \\
            \cmidrule{2-14}
            & \textbf{CIGPose-m} & 256$\times$192 & 2.3 & 59.9 & 69.4 & 69.0 & 76.6 & 64.3 & 77.6 & 82.1 & 88.3 & 49.7 & 61.6 \\
            & \textbf{CIGPose-l} & 256$\times$192 & 4.6 & 62.6 & 71.9 & 71.2 & 78.5 & 69.0 & 80.8 & 83.3 & 89.1 & 54.0 & 64.9 \\
            & \textbf{CIGPose-l} & 384$\times$288 & 10.7 & 66.3 & 74.9 & 73.0 & 79.9 & 72.0 & 83.3 & 88.3 & 92.3 & 59.8 & 70.0 \\
            & \textbf{CIGPose-x} & 384$\times$288 & 18.7 & 67.0 & 75.4 & 73.5 & 80.3 & 72.3 & 83.5 & 88.1 & 92.4 & 60.2 & 71.0 \\
            \cmidrule{2-14}
            & \textbf{CIGPose-l} + UBody & 384$\times$288 & 10.7 & 66.9 & 75.1 & 73.1 & 79.6 & 72.3 & 82.7 & 88.0 & 92.5 & 61.2 & 71.6 \\
            & \textbf{CIGPose-x} + UBody & 384$\times$288 & 18.7 & 67.5 & 75.5 & 73.5 & 80.1 & 70.3 & 81.6 & 88.4 & 92.7 & 62.6 & 72.6 \\
			\bottomrule
		\end{tabular}
	\end{center}
\end{table*}

\textbf{Datasets.} We conduct extensive experiments on four challenging benchmarks to validate our method. COCO \cite{lin2014microsoft} is a standard benchmark for 2D human pose estimation, for which we use the train2017 set for training and val2017 set for evaluation. COCO-WholeBody \cite{jin2020whole} extends COCO with dense annotations of 133 keypoints, enabling comprehensive whole-body analysis. UBody \cite{lin2023one} is a large-scale dataset featuring over 1M frames from 15 real-world scenarios. Following DWPose \cite{yang2023effective}, we sub-sample the videos at an interval of 10 frames for both training and validation. This dataset is used in conjunction with COCO-WholeBody to assess the generalization capability of our model. Finally, CrowdPose \cite{li2019crowdpose} is utilized to evaluate model performance in highly crowded and occluded scenes. We follow the standard protocol \cite{li2019crowdpose} of training on its trainval set and evaluating on the test set.

\textbf{Evaluation Metrics.} We follow the standard evaluation protocols for each dataset. For COCO \cite{lin2014microsoft} and COCO-WholeBody \cite{jin2020whole}, we report the standard OKS-based Average Precision (AP) and Average Recall (AR). For CrowdPose \cite{li2019crowdpose}, we report AP, AP$_{easy}$, AP$_{medium}$, and AP$_{hard}$ as defined in \cite{li2019crowdpose}.

\subsection{Implementation Details}
\label{sec:implement}

We implement CIGPose in the MMPose toolbox \cite{mmpose2020}, employing RTMPose \cite{jiang2023rtmpose} as the keypoint encoder with 256$\times$192 and 384$\times$288 inputs. To ensure a fair and reproducible comparison, all models are initialized with public RTMPose weights, which were pre-trained on COCO \cite{lin2014microsoft} and AIC \cite{wu2017ai} datasets. We train for up to 420 epochs using the AdamW optimizer (weight decay 0.05) and a cosine annealing learning rate schedule with a 1000-iteration linear warm-up. Our data augmentation follows RTMPose, including its two-stage fine-tuning approach. The weight $\lambda$ of counterfactual consistency loss $\mathcal{L}_{\text{cf}}$ is fixed at 0.1. During evaluation, we apply flip test and use person detection boxes consistent with RTMPose \cite{jiang2023rtmpose}, except for UBody \cite{lin2023one} where ground-truth boxes are used. Training is conducted on a single NVIDIA GeForce RTX 5090 for COCO \cite{lin2014microsoft} and CrowdPose \cite{li2019crowdpose} and 8 NVIDIA GeForce RTX 4090 GPUs for COCO-WholeBody \cite{jin2020whole} and UBody \cite{lin2023one}. We provide an expanded description of our network architecture and training settings in Appendix \ref{sec:appendix_implementation}, and our full two-stage data augmentation strategy in Appendix \ref{sec:appendix_augmentation}.

\subsection{Comparison with State-of-the-art Methods}

We evaluate our proposed CIGPose framework against state-of-the-art methods on three challenging public benchmarks: COCO-WholeBody \cite{jin2020whole}, COCO \cite{lin2014microsoft}, and CrowdPose \cite{li2019crowdpose}.

\begin{figure*}[!ht]
    \centering
    \includegraphics[width=\linewidth]{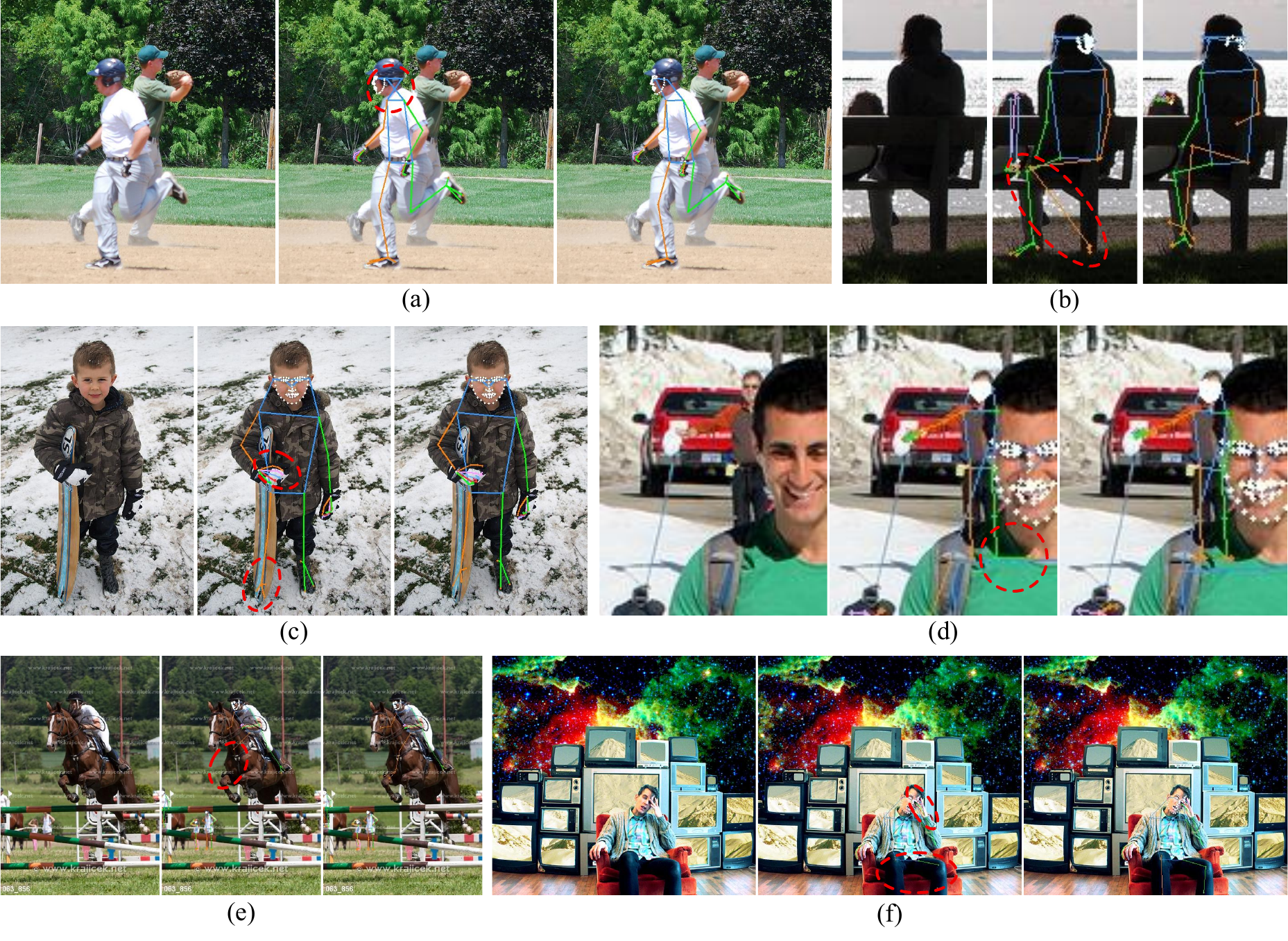}
    \caption{Qualitative comparison of CIGPose-x against the baseline RTMPose-x \cite{jiang2023rtmpose} on challenging images. From left to right: input image, RTMPose-x, and CIGPose-x.}
    \label{fig:contrast}
\end{figure*}

\textbf{COCO-WholeBody.} As shown in \cref{tab:compare_cocowhole}, CIGPose establishes a new SOTA on the COCO-WholeBody~\cite{jin2020whole} dataset. Trained solely on COCO-WholeBody, our CIGPose-x achieves 67.0\% AP, surpassing DWPose-l~\cite{yang2023effective} (66.5\% AP), which relies on two-stage distillation and additional data from UBody~\cite{lin2023one}. This highlights our framework's superior data efficiency and robustness. With the UBody dataset, CIGPose-x's performance further improves to 67.5\% AP. Notably, our CIGPose-l (384$\times$288) reaches 66.3\% AP, outperforming the larger RTMPose-x~\cite{jiang2023rtmpose} (65.3\% AP) with fewer GFLOPs. Qualitative results in \cref{fig:contrast} confirm our model's ability to produce more anatomically plausible poses in challenging scenes.

\begin{table}[htb]
  \centering
  \caption{Comparisons of CIGPose and SOTA methods on COCO val set \cite{lin2014microsoft}. The default input resolution is 256$\times$192, ``\dag'' denotes the input resolution is 384$\times$288.}
  \label{tab:cocovalResults}%
    \setlength{\tabcolsep}{0.02\linewidth}{\begin{tabular}{l|c|c|cccc}
    \toprule
    \multirow{2}[2]{*}{Method} & \multirow{2}[2]{*}{\#Params} & \multirow{2}[2]{*}{GFLOPs} & \multicolumn{2}{c}{COCO Val} \\
          &       &       & $AP$    & $AR$ \\
    \midrule
    SimpleBaseline~\cite{xiao2018simple}  & 60M  & 15.7  & 73.5  & 79.0 \\
    HRNet\dag~\cite{sun2019deep}  & 64M  & 32.9   & 76.3 & 81.2 \\
    HRFormer-B\dag~\cite{yuan2021hrformer}  & 43M  & 26.8  & 77.2  & 82.0 \\
    ViTPose++-B~\cite{xu2023vitpose++}  & 86M  & 17.1  & 77.0  & 82.6 \\
    RTMPose-m~\cite{jiang2023rtmpose}  & 14M  & 1.9  & 75.8  & 80.6 \\
    RTMPose-l~\cite{jiang2023rtmpose}  & 28M  & 4.2   & 76.5  & 81.3 \\
    RTMPose-l\dag~\cite{jiang2023rtmpose}  & 28M  & 9.3   & 77.3  & 81.9 \\
    \midrule
    CIGPose-m  & 14M   & 1.9  & 76.6  & 79.3 \\
    CIGPose-l  & 28M   & 4.2  & 77.6  & 80.3 \\
    CIGPose-l\dag  & 29M   & 9.4  & 78.5  & 81.1 \\
    \bottomrule
    \end{tabular}}%
\end{table}%

\textbf{COCO.} To demonstrate general applicability, we test on the COCO val2017 benchmark~\cite{lin2014microsoft} for 17-keypoint pose estimation. CIGPose consistently improves upon its baseline (\cref{tab:cocovalResults}). For instance, CIGPose-l with a 384$\times$288 input achieves 78.5\% AP, a 1.2 AP improvement over the strong RTMPose-l baseline with only a marginal increase in computational cost. This shows our causal mechanisms are beneficial for general pose estimation tasks.

\begin{table}[htb]
    \caption{Comparisons with SOTA methods on the CrowdPose \cite{li2019crowdpose} dataset. The default input resolution is 256$\times$192, ``\dag'' denotes the input resolution is 384$\times$288.}
    \label{tab:crowdpose}
    \centering
    \setlength{\tabcolsep}{3.3pt}
    \begin{tabular}{l|c|cccc}
        \toprule
        Method & \#Params & AP & AP$_{E}$ & AP$_{M}$ & AP$_{H}$ \\
        \midrule
        \multicolumn{6}{c}{Bottom-up methods} \\
        \midrule
        OpenPose~\cite{cao2019openpose} & - & - & 62.7 & 48.7 & 32.3 \\
        HrHRNet~\cite{cheng2020higherhrnet} & 63.8M & 65.9 & 73.3 & 66.5 & 57.9 \\
        DEKR~\cite{geng2021bottom} & 65.7M & 67.3 & 74.6 & 68.1 & 58.7 \\
        SWAHR~\cite{luo2021rethinking} & 63.8M & 71.6 & 78.9 & 72.4 & \textbf{63.0} \\
        \midrule
        \multicolumn{6}{c}{Top-down methods} \\
        \midrule
        SimpleBaseline~\cite{xiao2018simple} & 34.0M & 60.8 & 71.4 & 61.2 & 51.2 \\
        HRNet~\cite{sun2019deep} & 28.5M & 71.3 & 80.5 & 71.4 & 62.5 \\
        TransPose-H~\cite{yang2021transpose} & - & 71.8 & 79.5 & 72.9 & 62.2 \\
        HRFormer-B~\cite{yuan2021hrformer} & 43.2M & 72.4 & 80.0 & 73.5 & 62.4 \\
        RTMPose-m~\cite{jiang2023rtmpose} & 13.5M & 70.6 & 79.9 & 71.9 & 58.2 \\
        \midrule
        CIGPose-m & 14.4M & 71.4 & 81.0 & 72.7 & 58.9 \\
        CIGPose-l & 28.4M & \textbf{73.7} & \textbf{82.8} & \textbf{75.1} & 61.2 \\
        CIGPose-l\dag & 28.8M & 74.2 & 82.9 & 75.6 & 62.5 \\
        CIGPose-x\dag & 50.4M & 75.8 & 84.2 & 77.3 & 63.6 \\
        \bottomrule
    \end{tabular}
\end{table}

\textbf{CrowdPose.} We assess robustness in crowded and occluded scenarios using the CrowdPose~\cite{li2019crowdpose} dataset. \cref{tab:crowdpose} shows CIGPose achieves competitive results. Specifically, CIGPose-l obtains 73.7\% AP, outperforming prior SOTA methods like HRFormer-B~\cite{yuan2021hrformer} (72.4\% AP). A larger model, CIGPose-x\dag, further improves performance to 75.8\% AP. The consistent improvements on the medium and hard subsets underscore our method's efficacy in mitigating interference from common confounders like occlusions and cluttered backgrounds.

\subsection{Ablation Studies}
\label{sec:ablation_studies}

\begin{table}[!htb]
\caption{Ablation study of CIGPose on COCO-WholeBody. Values in parentheses are AP/AR gains.}
\label{tab:ablation_study}
\centering
\setlength{\tabcolsep}{4.3pt}
\begin{tabular}{ccc|c|cc}
\toprule 
CIM & $\mathcal{G}_h$ & $\mathcal{G}_p$ & Model & AP & AR \\
\midrule
\multirow{2}{*}{\cmark} & \multirow{2}{*}{\cmark} & \multirow{2}{*}{\cmark} & CIGPose-l & 66.3 ({+1.5}) & 74.9 ({+1.9}) \\
                            &                             &                             & CIGPose-x & 67.0 ({+1.7}) & 75.4 ({+2.1}) \\
\midrule
\multirow{2}{*}{\xmark} & \multirow{2}{*}{\cmark} & \multirow{2}{*}{\cmark} & CIGPose-l     & 66.0 ({+1.2}) & 74.8 ({+1.8}) \\
                            &                             &                             & CIGPose-x     & 66.8 ({+1.5}) & 75.5 ({+2.2}) \\
\midrule
\multirow{2}{*}{\xmark} & \multirow{2}{*}{\xmark} & \multirow{2}{*}{\cmark} & CIGPose-l     & 65.6 ({+0.8}) & 74.6 ({+1.6}) \\
                            &                             &                             & CIGPose-x     & 66.3 ({+1.0}) & 75.0 ({+1.7}) \\
\midrule
\multirow{2}{*}{\cmark} & \multirow{2}{*}{\xmark} & \multirow{2}{*}{\xmark} & CIGPose-l     & 65.7 ({+0.9}) & 74.3 ({+1.3}) \\
                            &                             &                             & CIGPose-x     & 66.1 ({+0.8}) & 74.9 ({+1.6}) \\
\midrule
\multirow{2}{*}{\xmark} & \multirow{2}{*}{\xmark} & \multirow{2}{*}{\xmark} & \multirow{2}{*}{Baseline~\cite{jiang2023rtmpose}}     & 64.8 & 73.0 \\
                            &                             &                  &                             & 65.3 & 73.3 \\
\bottomrule
\end{tabular}
\end{table}

We conduct ablation studies on the COCO-WholeBody \cite{jin2020whole} dataset to dissect the contributions of key components in CIGPose. All models use RTMPose \cite{jiang2023rtmpose} as keypoint encoder and a 384$\times$288 input size. 

\textbf{Effectiveness of Proposed Components.} As shown in \cref{tab:ablation_study}, we analyze the individual contributions of the Causal Intervention Module (CIM) and the hierarchical GNN components ($\mathcal{G}_p$ and $\mathcal{G}_h$) against the baseline \cite{jiang2023rtmpose}. Adding the hierarchical GNN components ($\mathcal{G}_p$ and $\mathcal{G}_h$) without intervention yields 66.8 AP, demonstrating the value of structural reasoning (+1.5 AP). Introducing our CIM on top of the full GNN further lifts performance to 67.0 AP, a +0.2 AP gain from deconfounding. The full CIGPose-x model achieves a total gain of 1.7 AP over the baseline, highlighting the synergistic effect of performing hierarchical reasoning on deconfounded embeddings. We provide further ablation studies on intervention frequency, the consistency loss weight $\lambda$, and alternative intervention strategies in Appendix \ref{sec:appendix_extra_exp}.
\section{Conclusion}

We introduced CIGPose, a novel framework that applies causal intervention to whole-body pose estimation. We proposed using predictive uncertainty as a proxy for confounding and quantitatively validated its effectiveness in identifying heavy occlusion, a critical and severe class of confounders. By replacing these keypoints with learned canonical embeddings, our method mitigates spurious correlations and achieves state-of-the-art results on major 2D benchmarks. Future work will focus on extending our causal framework to more challenging domains, such as 3D and out-of-distribution scenarios. We provide a discussion of the current limitations of our approach, including the potential risks of our intervention strategy and the constraints of our confounding proxy, in Appendix \ref{sec:appendix_limitations}. This approach opens a promising direction for building more reliable and generalizable pose estimators.

\paragraph{Acknowledgement.}This work was supported by the Key Research and Development Program of Shaanxi (2024GX-YBXM-543), the Fundamental Research Funds for the Central Universities (QTZX25114), Zhejiang Provincial Excellent Funding for Postdoctoral Researchers (ZJ2025156) and China Postdoctoral Science Foundation (2025M784428).
{
    \small
    \bibliographystyle{ieeenat_fullname}
    \bibliography{main}
}

\clearpage
\setcounter{page}{1}
\maketitlesupplementary

\appendix

\section{Theoretical Foundation of Causal Intervention}
\label{sec:appendix_theory}

This section provides the detailed theoretical support for the causal intervention framework introduced in the main paper \cref{sec:method}. We first provide a formal derivation for the backdoor adjustment formula \cref{eq:backdoor_adjustment} using Pearl's $do$-calculus \cite{pearl2016causal} and then offer a theoretical justification for our counterfactual replacement approximation \cref{eq:cf_embedding_replace}.

\myparagraph{Derivation of the Backdoor Adjustment Formula.}

Our goal is to estimate the interventional distribution $P(Y|do(F))$, which represents the true causal effect of the keypoint embeddings $F$ on the final prediction $Y$, isolated from the confounding influence of visual context $C$.

We recall the Structural Causal Model (SCM) from \cref{fig:cim}(a) of the main paper. This model defines the causal paths $C \to X \to F \to Y$ and the confounding paths $C \to X$ and $C \to Y$. The critical issue is the non-causal backdoor path $F \leftarrow X \leftarrow C \to Y$, which allows spurious correlations between $F$ and $Y$ based on the confounder $C$.

To block this path, we must adjust for the confounding variable $C$. Using the three rules of $do$-calculus \cite{pearl2016causal}, we can formally derive the backdoor adjustment formula presented in \cref{eq:backdoor_adjustment}.

Given a causal graph $\mathcal{G}$, $\mathcal{G}_{\overline{X}}$ denotes the graph with incoming edges to $X$ removed, and $\mathcal{G}_{\underline{X}}$ denotes the graph with outgoing edges from $X$ removed.

\begin{itemize}
    \item \textbf{Rule 1 (Insertion/deletion of observations):}
    \begin{equation*}
    \begin{split}
        P(y|do(x), z, w) = P(y|do(x), w), \\
        \qquad \text{ if } (Y \perp Z | X, W)_{\mathcal{G}_{\overline{X}}}.
    \end{split}
    \end{equation*}
    
    \item \textbf{Rule 2 (Action/observation exchange):}
    \begin{equation*}
    \begin{split}
        P(y|do(x), z, w) = P(y|x, z, w), \\
        \qquad \text{ if } (Y \perp X | Z, W)_{\mathcal{G}_{\underline{X}}}.
    \end{split}
    \end{equation*}
    
    \item \textbf{Rule 3 (Insertion/deletion of actions):}
    \begin{equation*}
    \begin{split}
        P(y|do(x), do(z), w) = P(y|do(x), w), \\
        \qquad \text{ if } (Y \perp Z | X, W)_{\mathcal{G}_{\overline{X}, \overline{Z(W)}}}.
    \end{split}
    \end{equation*}
\end{itemize}

Our derivation proceeds as follows:
\begin{align*}
P(Y|do(F)) & = \sum_{c} P(Y|do(F), c) P(c|do(F)) \label{eq:a1} \tag{A1} \\
& = \sum_{c} P(Y|do(F), c) P(c) \label{eq:a2} \tag{A2} \\
& = \sum_{c} P(Y|F, c) P(c) \label{eq:a3} \tag{A3}
\end{align*}

\textbf{Step \ref{eq:a1}:} We begin by applying the law of total probability, marginalizing over the confounder $C$.

\textbf{Step \ref{eq:a2}:} We apply \textbf{Rule 3} (Insertion/deletion of actions) to show $P(c|do(F)) = P(c)$. This holds if $(C \perp F | \emptyset)_{\mathcal{G}_{\overline{F}}}$. In the graph $\mathcal{G}_{\overline{F}}$, the edge $X \to F$ is removed. The only paths from $C$ to $F$ are $C \to X \quad F$ (d-separated) and $C \to Y \leftarrow F$. The latter is a v-structure and is blocked since $Y$ is not a condition. Thus, $C$ and $F$ are d-separated, $(C \perp F | \emptyset)_{\mathcal{G}_{\overline{F}}}$ holds, and the intervention $do(F)$ has no effect on $C$.

\textbf{Step \ref{eq:a3}:} We apply \textbf{Rule 2} (Action/observation exchange) to show $P(Y|do(F), c) = P(Y|F, c)$. This holds if $(Y \perp F | C)_{\mathcal{G}_{\underline{F}}}$. In the graph $\mathcal{G}_{\underline{F}}$, the edge $F \to Y$ is removed. We must check if any other paths connect $F$ and $Y$, given $C$. The only other path is the backdoor path $F \leftarrow X \leftarrow C \to Y$. Since we are conditioning on $C$, this path is blocked at $C$. Therefore, $(Y \perp F | C)_{\mathcal{G}_{\underline{F}}}$ holds, and we can replace the action $do(F)$ with the observation $F$.

This completes the formal derivation of \cref{eq:backdoor_adjustment} in the main paper.

\myparagraph{Justification for Counterfactual Replacement.}

As stated in the main text, \cref{eq:a3} (the backdoor adjustment formula) is intractable to compute because $C$ is unobserved, high-dimensional, and impossible to sum over. Our Causal Intervention Module (CIM) approximates this intervention by performing a \textit{counterfactual replacement}, $do(f_k := z_k)$, as defined in \cref{eq:cf_embedding_replace}.

The theoretical justification is as follows:
\begin{enumerate}
    \item \textbf{Goal:} The target distribution $P(Y|do(F)) = \mathbb{E}_{c}[P(Y|F, c)]$ represents the causal effect of $F$ on $Y$, averaged over all possible contexts $c$. This conceptually creates a context-invariant, unbiased representation.

    \item \textbf{Problem:} The observed embedding $f_k \in F$ is confounded. It is a descendant of $C$ via the path $C \to X \to F$. This dependency on $C$ is precisely what opens the spurious backdoor path $F \leftarrow X \leftarrow C \to Y$.

    \item \textbf{Our Approximation:} We introduce a learnable canonical embedding table $Z \in \mathbb{R}^{K \times d_{\mathit{emb}}}$. As a shared global parameter matrix, $Z$ is reused across all images and is not conditioned on the current instance. Hence, it is optimized end-to-end while remaining independent of any specific input image $X_i$ and its associated confounder $C_i$; that is, $Z \perp C$ by construction.

    \item \textbf{The Intervention:} When our CIM identifies a confounded embedding $f_k$ (via $s_c(k)$) and replaces it with its corresponding canonical ideal $z_k \in Z$, it is performing the counterfactual operation $do(f_k := z_k)$. This operation replaces a variable that is dependent on $C$ (the original $f_k$) with a variable that is independent of $C$ (the canonical $z_k$).

    \item \textbf{The Effect:} This replacement physically severs the causal link $C \to X \to F$ at the feature level. For the intervened embedding $f'_k = z_k$, the backdoor path $F' \leftarrow X \leftarrow C \to Y$ is broken because $f'_k$ is no longer a descendant of $C$ or $X$.
\end{enumerate}
The shared/global parameterization of $Z$ is central to this approximation. Each row $z_k$ accumulates updates from many images whenever keypoint type $k$ is selected for intervention, so sample-specific contextual cues are not tied to any single replacement vector. Instead, optimization reinforces what is consistently useful for predicting the same anatomical keypoint across diverse contexts, which is precisely the intended context-invariant prior. By computing $P(Y|F')$ where $F'$ is the deconfounded set, we force the model to reason using the context-invariant ideal $z_k$ instead of the confounded evidence $f_k$. This serves as a practical, sample-specific approximation of the full, intractable summation over all contexts $\sum_{c}$. The counterfactual consistency loss defined in \cref{eq:loss_cf} further ensures that this replacement is targeted, regularizing $Z$ to be a meaningful "ideal" representation for stable keypoints.

\section{Further Analysis of Predictive Uncertainty}
\label{sec:appendix_uncertainty_viz}

As discussed in the main paper \cref{sec:intro} and \cref{sec:method}, our framework posits that predictive uncertainty is an effective proxy for identifying confounded keypoint representations. Confounders like heavy occlusion create a conflict between the model's spuriously learned priors from $C \to Y$ and the visual evidence from $F \to Y$, resulting in high epistemic uncertainty.

\myparagraph{Beyond-occlusion quantitative evidence.}
The occlusion-based analysis in \cref{fig:confounder_boxplot} validates $s_c(k)$ against a clearly defined confounder. To test whether the proxy also captures broader keypoint difficulty beyond explicit occlusion labels, we perform a within-instance top-$n$ enrichment analysis on the COCO-WholeBody validation set. For instance $i$, let $V_i$ be the set of visible keypoints and let $T_i = \mathrm{TopK}(\{s_{c,i}(k)\}, n)$ be the keypoints selected by the proxy. We define
\begin{equation}
\label{eq:topn_enrichment}
    \begin{aligned}
        \Delta_i &= \frac{1}{|T_i|}\sum_{k\in T_i} e_i(k) - \frac{1}{|V_i \setminus T_i|}\sum_{k\in V_i\setminus T_i} e_i(k), \\
        e_i(k) &= \left\| \hat{\mathbf{p}}_i(k) - \mathbf{p}_i(k) \right\|_2,
    \end{aligned}
\end{equation}
where $e_i(k)$ is the localization error of keypoint $k$. A positive $\Delta_i$ means that, within the same instance, the keypoints selected by $s_c$ incur larger error than the remaining visible keypoints.

\begin{table}[t]
\centering
\caption{Within instance top-$n$ enrichment on COCO-WholeBody. Keypoints selected by $s_c$ exhibit larger localization error than the remaining keypoints in the same instance. Easy-drop $p$ denotes the fraction of easiest instances removed when forming harder subsets, and 95\% CIs are percentile bootstrap intervals over instances.}
\label{tab:topn_enrichment}
\small
\setlength{\tabcolsep}{4pt}
\begin{tabular}{lccc}
\toprule
Easy-drop $p$ & Kept inst. & Mean $\bar{\Delta}$ (px) & 95\% CI \\
\midrule
0.00 & 104,125 & 8.47 & [8.20, 8.72] \\
0.30 & 72,888 & 10.14 & [9.78, 10.48] \\
0.50 & 52,063 & 10.45 & [10.02, 10.87] \\
\bottomrule
\end{tabular}
\end{table}

As shown in \cref{tab:topn_enrichment}, the enrichment is strongly positive and widens on progressively harder subsets. This indicates that $s_c$ systematically concentrates on intrinsically harder keypoints within an instance, including cases driven by clutter, blur, and truncation rather than only binary occlusion status.

\begin{figure}[h]
  \centering
  \includegraphics[width=1.0\linewidth]{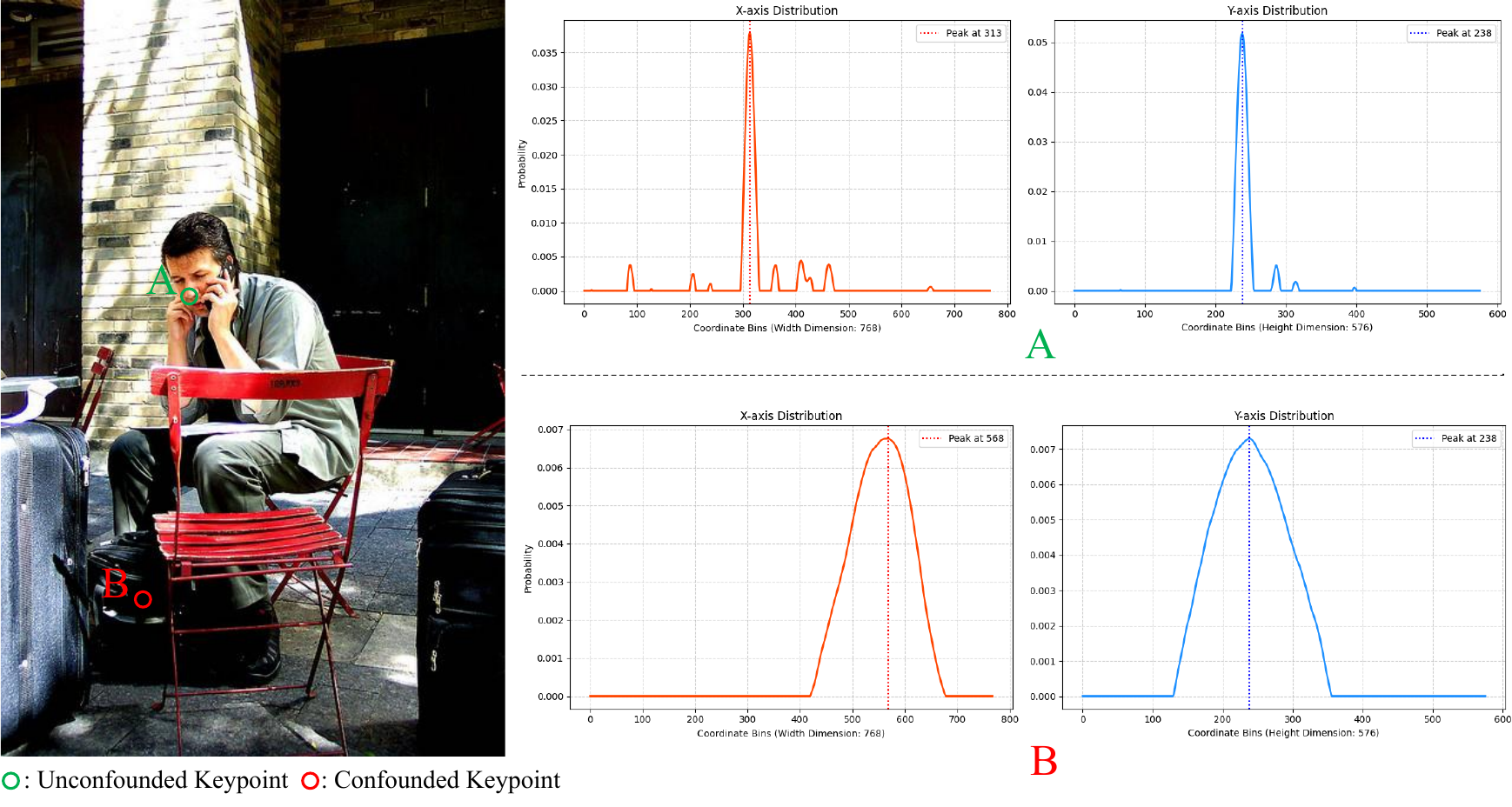}
  \caption{Visualization of posterior probability distributions. (A) For an unconfounded keypoint (green circle, nose), the distributions exhibit sharp, high peaks, indicating low uncertainty. (B) For a confounded keypoint (red circle, left ankle), which is occluded and in shadow, the distributions are diffuse with low peaks, signaling high predictive ambiguity.}
  \label{fig:uncertainty_viz_moved}
\end{figure}

\myparagraph{Qualitative example.}
In \cref{fig:confounder_boxplot} of the main text, we quantitatively validated the proxy with occlusion labels. Here, in \cref{fig:uncertainty_viz_moved}, we provide the qualitative example referenced in the main text.

This example visually demonstrates the mechanism of CIGPose:
\begin{enumerate}
    \item \textbf{Confounder Identification:} The occluded L-Ankle leads to high predictive uncertainty, producing a diffuse posterior distribution.
    \item \textbf{Causal Intervention:} Our model computes a high $s_c(k)$ for this keypoint, triggering the counterfactual replacement $do(f_{L\text{-}Ankle} := z_{L\text{-}Ankle})$.
    \item \textbf{Deconfounded Reasoning:} The Hierarchical GNN, now operating on the "clean" embedding set $F'$, leverages anatomical constraints from visible keypoints (such as the L-Knee) to infer the position of the occluded joint, resulting in an anatomically plausible prediction.
\end{enumerate}

\section{Expanded Implementation Details}
\label{sec:appendix_implementation}

This section provides additional details on network architecture and training settings, supplementing \cref{sec:method} and \cref{sec:implement} of the main paper.

\myparagraph{Network Architecture Details.}
Our CIGPose framework builds upon the RTMPose \cite{jiang2023rtmpose} architecture, introducing a novel prediction head. Below, we detail the components of this head.

\begin{itemize}
    \item \textbf{Backbone and Feature Processing}: We employ a CSPNeXt backbone as our keypoint encoder. For an input image of size $H \times W$, the backbone outputs a feature map of size $\frac{H}{32} \times \frac{W}{32}$ with 1280 channels. This feature map is then processed by a Gated Attention Unit (GAU) with hidden dimensions of 512, which extracts the initial keypoint embeddings $F \in \mathbb{R}^{B \times K \times 512}$, where $B$ is the batch size and $K$ is the number of keypoints ($K = 133$ for the COCO-WholeBody dataset, and $K = 17$ for the COCO and CrowdPose datasets).

    \item \textbf{Causal Intervention Module (CIM)}: The CIM takes the initial embeddings $F$ as input. First, two separate linear layers project $F$ to generate initial 1D SimCC predictions $(P_{k,x}, P_{k,y})$. These are used to calculate the confounder score $s_c(k)$ for each keypoint per \cref{eq:confounder_identification}. The module then identifies the top-$n$ most confounded embeddings and replaces them with corresponding vectors from a shared canonical embedding table $Z \in \mathbb{R}^{K \times 512}$. This table is implemented as a standard \texttt{Embedding} layer, initialized with a normal distribution $\mathcal{N}(0, 0.01^2)$ and optimized end-to-end.

    \item \textbf{Hierarchical GNN}: The deconfounded embeddings $F'$ are processed by a hierarchical GNN to model anatomical constraints, as discussed in \cref{sec:hierarchical_gnn}.
    \begin{itemize}
        \item \textit{Intra-Part Modeling}: The first stage uses an \texttt{EdgeConv} \cite{wang2019dynamic} layer that operates over the standard anatomical skeleton graph $\mathcal{G}_p$. This layer models local kinematics by computing edge features (the difference between connected node features concatenated with the source node feature) and aggregating them via a shared MLP, which is implemented using a 1$\times$1 2D convolution.
        \item \textit{Inter-Part Attention}: The second stage uses an \texttt{AttentionModule} to capture long-range dependencies. It first computes an aggregate representation for each predefined semantic keypoint group by averaging the features of its constituent keypoints. These group-level features are then refined by passing them through a second \texttt{EdgeConv} layer defined over a fully-connected graph of all groups. The output is then used to generate channel-wise attention weights via a linear layer and a sigmoid function, per \cref{eq:attention}. These weights modulate the keypoint features, enabling the model to reason about global inter-part relationships. The full skeleton graph definition and the list of semantic groups are specified in the model configuration files.
    \end{itemize}
\end{itemize}

\myparagraph{Joint optimization of the canonical embedding table.}
For a mini-batch of size $B$, the counterfactual replacement in \cref{eq:cf_embedding_replace} is applied instance-wise as
\begin{equation}
\label{eq:batch_replace}
\begin{aligned}
f'_{b,k} &= (1-m_{b,k}) f_{b,k} + m_{b,k} z_k, \\
m_{b,k} &= \mathbf{1}[k \in \mathcal{I}_b],
\end{aligned}
\end{equation}
where $\mathcal{I}_b = \mathrm{TopK}(\{s_c(b,k)\}_{k=1}^{K}, n)$ is the selected set for instance $b$, and the mask $M=\{m_{b,k}\}$ is broadcast along the embedding dimension. Let $H_{\theta}(\cdot)$ denote the prediction head following the encoder and hierarchical graph reasoning. The counterfactual path computes $P_{\text{cf}}(Y)=H_{\theta}(F')$, while the observational prediction $P_{\text{obs}}(Y)=H_{\theta}(F)$ is used only through the stop-gradient target in \cref{eq:loss_cf}. Therefore, gradients flow into the selected rows of $Z$ according to
\begin{equation}
\label{eq:z_gradient}
\frac{\partial \mathcal{L}}{\partial z_k} = \sum_{b=1}^{B} m_{b,k} \frac{\partial \mathcal{L}}{\partial f'_{b,k}},
\end{equation}
while rows that are not selected in the current mini-batch receive zero gradient from that step. Thus, $Z$ is optimized jointly with the network parameters $\theta$ under $\mathcal{L}=\mathcal{L}_{\text{kpt}}+\lambda\mathcal{L}_{\text{cf}}$.

The corresponding training procedure is summarized in \cref{alg:optZ}.

\begin{algorithm}[htbp]
\caption{Training: joint update of $(\theta,Z)$}
\label{alg:optZ}
\footnotesize
\DontPrintSemicolon
\KwIn{mini-batch images $X$; GT distributions $\{Q_{b,k}\}$; budget $n$; weight $\lambda$}
\KwOut{updated parameters $(\theta,Z)$}
$F \leftarrow E_\theta(X)$\tcp*{keypoint embeddings}
Compute $(P_{b,k,x},P_{b,k,y})$ and scores $s_c(b,k)$ (Eq.~(2))\;
\For{$b=1$ \KwTo $B$}{
  $\mathcal{I}_b \leftarrow \mathrm{TopK}(\{s_c(b,k)\}_{k=1}^{K}, n)$\;
  $m_{b,k}\leftarrow \mathbf{1}[k\in\mathcal{I}_b]\ ,\forall k\in\{1,\ldots,K\}$\;
}
$F' \leftarrow (1-M)\odot F + M\odot Z$\tcp*{Eq.~(3), $M=\{m_{b,k}\}$}
$P_{\text{cf}} \leftarrow H_\theta(F')$\;
$\tilde P_{\text{obs}} \leftarrow \mathrm{sg}[H_\theta(F)]$\tcp*{stop-grad target}
$\mathcal{L}_{\text{kpt}} \leftarrow \sum_{b,k} w_{b,k}\,D_{\mathrm{KL}}(Q_{b,k}\,\|\,P_{\text{cf}}(Y_{b,k}))$\;
$\mathcal{L}_{\text{cf}} \leftarrow \frac{1}{\sum_b|S_b|}\sum_b\sum_{k\in S_b}
D_{\mathrm{KL}}(\tilde P_{\text{obs}}(Y_{b,k})\,\|\,P_{\text{cf}}(Y_{b,k}))$\tcp*{$S_b$: stable (non-intervened) set}
$\mathcal{L}\leftarrow \mathcal{L}_{\text{kpt}}+\lambda\mathcal{L}_{\text{cf}}$; update $(\theta,Z)$ by AdamW\;
\end{algorithm}

\myparagraph{Training settings.}
Our training settings largely follow that of RTMPose \cite{jiang2023rtmpose}, including a two-stage fine-tuning approach. The key hyperparameters for our main experiments (CIGPose-x on COCO-WholeBody) are summarized in \cref{tab:hyperparams}.

\begin{table}[h]
\centering
\caption{Key training hyperparameters for the CIGPose-x model.}
\label{tab:hyperparams}
\setlength{\tabcolsep}{12pt}
\begin{tabular}{@{}ll@{}}
\toprule
\textbf{Hyperparameter} & \textbf{Value} \\ \midrule
Optimizer & AdamW \\
Base Learning Rate & $2 \times 10^{-3}$ \\
Weight Decay & 0.05 \\
LR Schedule & Cosine Annealing \\
Warm-up Iterations & 1000 \\
Max Gradient Norm & 35 \\
Training Epochs & 420 \\
Stage-2 Epochs & 150 \\
$\lambda$ for $\mathcal{L}_{cf}$ & 0.1 (Ablated in \cref{sec:appendix_extra_exp}) \\
Train Batch Size & 32 (per GPU) \\
Intervention $n$ & 13 (Ablated in \cref{sec:appendix_extra_exp}) \\ \bottomrule
\end{tabular}
\end{table}

\section{Data Augmentation}
\label{sec:appendix_augmentation}

We adopt a two-stage data augmentation strategy to enhance model robustness, corresponding to the two-stage training detailed in \cref{sec:appendix_implementation}.
\begin{itemize}
    \item \textbf{Stage 1 (Epoch 1-270)}: This stage uses aggressive augmentation. It includes random horizontal flipping, random half-body transforms, random bounding box transformations (scaling from 0.5x to 1.5x, rotation of $\pm 90^\circ$), top-down affine transforms, and YOLOX-style HSV color jittering. To simulate heavy occlusions, we also apply Albumentation transforms including \texttt{Blur} (p=0.1), \texttt{MedianBlur} (p=0.1), and \texttt{CoarseDropout} (p=1.0, max\_holes=1, max\_height=128, max\_width=128).
    \item \textbf{Stage 2 (Epoch 271-420)}: For the final fine-tuning stage, the augmentation intensity is reduced to allow the model to converge on cleaner data. Specifically, the random bounding box transform is modified to perform only scaling and rotation (shift factor is set to 0), and the probability of applying \texttt{CoarseDropout} is lowered to 0.5.
\end{itemize}

\section{Additional Experimental Analysis}
\label{sec:appendix_extra_exp}

We conduct further experiments on the COCO-WholeBody \cite{jin2020whole} validation set to analyze key components of our framework, supplementing the main ablations in \cref{sec:ablation_studies}. Unless otherwise specified, these studies use the CIGPose-l (384$\times$288) model.

\myparagraph{Analysis of Intervention Frequency.}
To verify that our Causal Intervention Module effectively targets keypoints susceptible to confounding, we analyze the intervention frequency for different body parts on the COCO-WholeBody validation set. As shown in \cref{tab:intervention_freq}, the intervention rates are highest for feet, hands, and legs. These are the body parts most frequently affected by common visual confounders such as occlusion, motion blur, and truncation. This empirical result provides evidence that our uncertainty-based proxy successfully identifies and targets the most confounded keypoints.

\begin{table}[htb]
\centering
\caption{Intervention frequency per body part on COCO-WholeBody. The module intervenes most often on extremities, which are most prone to confounding.}
\label{tab:intervention_freq}
\setlength{\tabcolsep}{15pt}
\begin{tabular}{@{}lc@{}}
\toprule
\textbf{Body Part} & \textbf{Intervention Freq. (\%)} \\ \midrule
Face & 0.67 \\
Torso & 0.16 \\
Arms & 0.24 \\
Hands & 0.90 \\
Legs & 0.89 \\
Feet & 1.36 \\ \bottomrule
\end{tabular}
\end{table}

\myparagraph{Sensitivity to Consistency Loss Weight $\lambda$.}
The counterfactual consistency loss, $\mathcal{L}_{cf}$ which is defined in \cref{eq:loss_cf}, is critical for ensuring that interventions are meaningful and that the canonical embeddings $Z$ are learned effectively. We analyze the model's performance while varying the loss weight $\lambda$.

\begin{table}[htb]
\centering
\caption{Effect of the counterfactual consistency loss weight $\lambda$.}
\label{tab:ablation_lambda}
\setlength{\tabcolsep}{15pt}
\begin{tabular}{@{}lc@{}}
\toprule
\textbf{$\lambda$} & \textbf{Whole-Body AP (\%)} \\ \midrule
0 & 65.8 \\
0.01 & 66.1 \\
\textbf{0.1} & \textbf{66.3} \\
0.5 & 65.9 \\ \bottomrule
\end{tabular}
\end{table}

As shown in \cref{tab:ablation_lambda}, removing the consistency loss entirely ($\lambda=0$) leads to a 0.5 AP drop. This demonstrates that without this regularization, the model struggles to learn stable and meaningful canonical embeddings. A larger weight ($\lambda=0.5$) also slightly degrades performance, likely by overly constraining the GNN and preventing it from fully leveraging the deconfounded features. Our chosen value of $\lambda=0.1$ provides the best trade-off.

\myparagraph{Analysis of Intervention Strategy and Parameter $n$.}
All main experiments use a fixed `top-$n$` intervention strategy during training. For comparison, we also evaluate a threshold-based strategy. Formally, given a predefined threshold $\tau$, the threshold strategy defines the set of keypoints selected for intervention, $\mathcal{K}_{\text{conf}}$, as:
$$
\mathcal{K}_{\text{conf}} = \{k \mid s_c(k) > \tau\}
$$
where $s_c(k)$ is the confounder score defined in \cref{eq:confounder_identification}. This means we intervene on any keypoint whose uncertainty-based score surpasses the threshold.

\begin{table}[htb]
\centering
\caption{Comparison of intervention strategies during training. The fixed-budget `top-$n$` strategy provides a more stable and effective training signal, with $n=13$ yielding the best result on COCO-WholeBody.}
\label{tab:ablation_strategy}
\small
\begin{tabular}{@{}lc@{}}
\toprule
\textbf{Intervention Strategy (Training)} & \textbf{Whole-Body AP (\%)} \\ \midrule
Threshold $\tau = 0.7$ & 65.9 \\
Threshold $\tau = 0.8$ & 65.8 \\
\midrule
Top-$n$ ($n=11$) & 66.1 \\
\textbf{Top-$n$ ($n=13$)} & \textbf{66.3} \\
Top-$n$ ($n=15$) & 66.0 \\ \bottomrule
\end{tabular}
\end{table}

The results in \cref{tab:ablation_strategy} show that the `top-$n$` strategy is more effective. We hypothesize that this is because it provides a more stable training signal by ensuring a fixed number of keypoints $n$ are intervened upon in each iteration. The `threshold` approach can be noisy, as the number of interventions can vary dramatically between samples, potentially making it more difficult for the model to learn effective canonical embeddings. We also ablate the value of $n$ (for training) and find $n=13$ (10\% of $K=133$ keypoints) to be optimal.

\section{Limitations}
\label{sec:appendix_limitations}

\begin{figure}[h]
  \centering
  \includegraphics[width=1.0\linewidth]{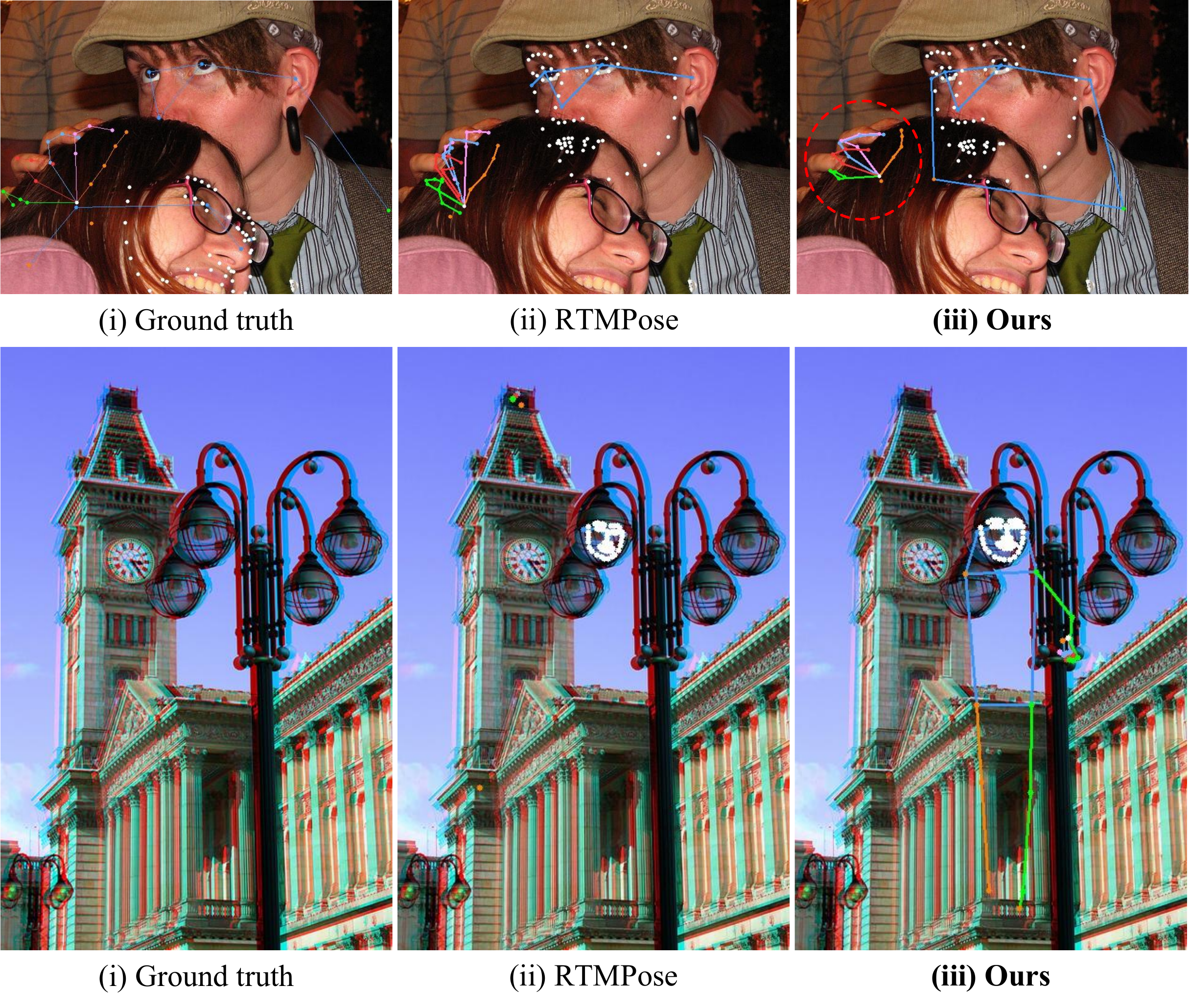}
  \caption{Visualization of failure cases.}
  \label{fig:limitations_viz}
\end{figure}

\myparagraph{Limitations of Causal Intervention via Replacement.}
The Causal Intervention Module (CIM) approximates the intractable $do$-operation by identifying confounded embeddings $f_k$, as signaled by high uncertainty $s_c(k)$, and replacing them with context-invariant canonical embeddings $z_k$. While effective for occlusion, this $do(f_k := z_k)$ substitution carries a risk of over regularization. The core assumption is that high uncertainty correlates with visual confounding. However, this assumption can be violated in cases of valid but statistically rare poses, such as the intimate head-to-head interaction shown in \cref{fig:limitations_viz} (top). In such scenarios, the visual evidence $f_k$ is anatomically correct but lies on the tail of the pose distribution. The model exhibits high epistemic uncertainty simply due to the sample's out-of-distribution nature. Consequently, CIM may incorrectly flag these valid embeddings as confounded and replace them with $z_k$, which represents a mean or canonical ideal. This can regularize the prediction too aggressively, discarding correct fine-grained geometry in favor of a generic high-likelihood pose.

\myparagraph{Limitations of Uncertainty as a Confounding Proxy.}
The second limitation relates to the proxy $s_c(k)$ itself. Our quantitative validation in \cref{fig:confounder_boxplot,tab:topn_enrichment} shows that $s_c(k)$ is effective at identifying ambiguity-induced difficulty, including occlusion and harder non-occlusion cases. However, it fails when the model falls victim to confident semantic errors. As illustrated in \cref{fig:limitations_viz} (bottom), the model hallucinates a human skeleton on a street lamp. This error stems from spurious correlations learned during training---specifically, associating vertical linear structures with limbs. Critically, the model does not view this as an ambiguous state; it predicts the false positive with high confidence (low uncertainty), likely due to the strong shape similarity fulfilling the top-down detector's prior. Because $s_c(k)$ remains low, the CIM mechanism is bypassed entirely. This highlights a fundamental boundary of our framework: it is adept at rectifying uncertain representations but lacks an inherent rejection mechanism for confident misidentification caused by strong contextual confusion.

\clearpage
\onecolumn

\section{More Qualitative Results}
\label{sec:appendix_qualitatives}

\vspace*{\fill}
\begin{center}
    \includegraphics[width=\textwidth]{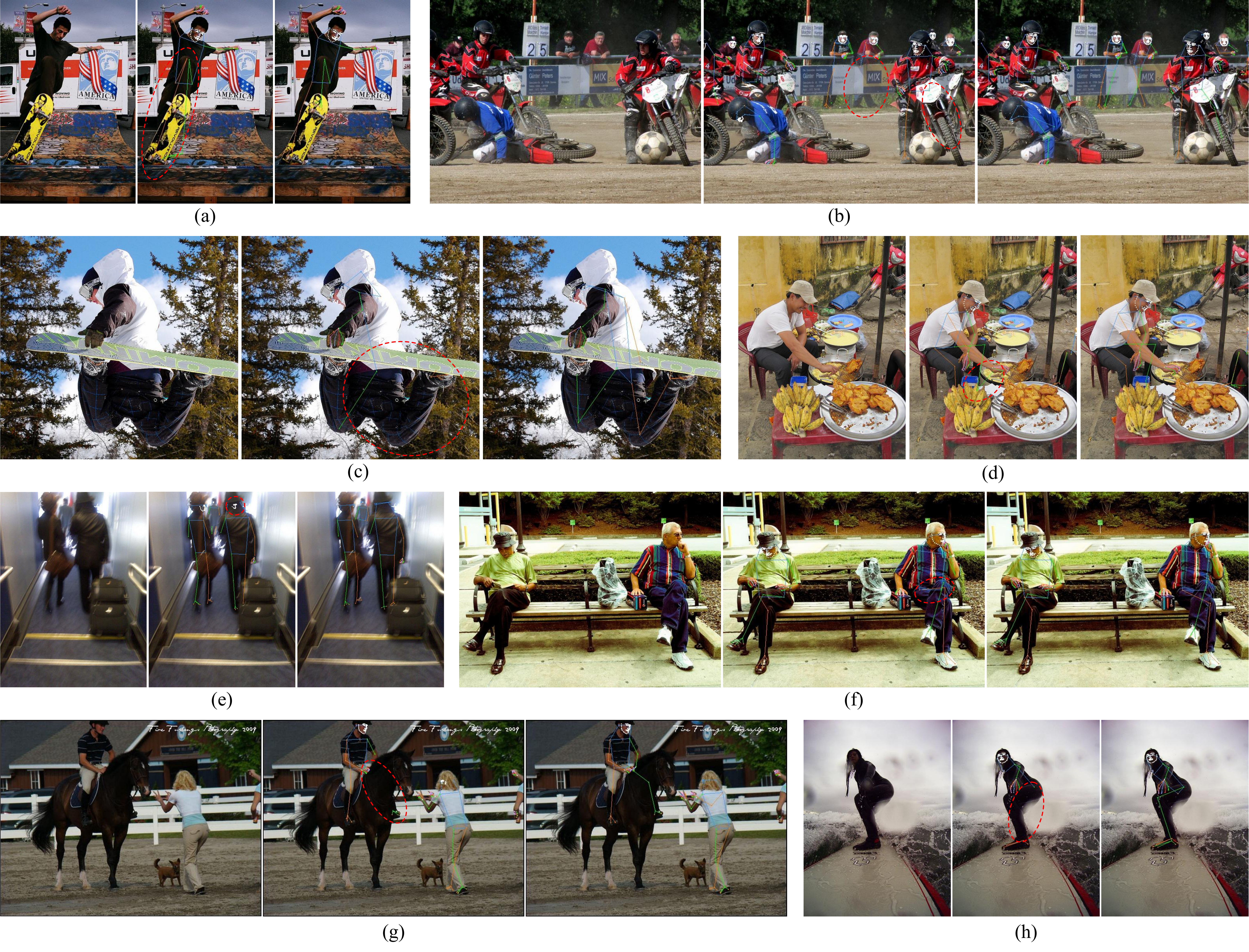}
    
    \captionof{figure}{Qualitative comparison of CIGPose-x against the baseline RTMPose-x \cite{jiang2023rtmpose} on challenging images. From left to right: input image, RTMPose-x, and CIGPose-x.}
    \label{fig:more_contrast}
\end{center}
\vspace*{\fill}

\end{document}